\pdfoutput=1

\documentclass[11pt]{article}

\usepackage{authblk}

\usepackage[]{acl}

\usepackage{times}
\usepackage{latexsym}
\usepackage{todonotes}
\usepackage{multirow}
\usepackage{stfloats}

\usepackage[T1]{fontenc}

\usepackage[utf8]{inputenc}

\usepackage{microtype}

\newcommand{\kpx}{KPE}
\newcommand{\mdkpx}{M\kpx}
\newcommand{\sect}{\S}
\newcommand{\dname}{MK-DUC-01}

\title{Multi-Document Keyphrase Extraction: \\Dataset, Baselines and Review}

\author[1]{\bf Ori Shapira}
\author[2]{\bf Ramakanth Pasunuru}
\author[1]{\bf Ido Dagan}
\author[1]{\bf Yael Amsterdamer}
{
\makeatletter
\renewcommand\AB@affilsepx{~~~ \protect\Affilfont} \makeatother
\affil[1]{Bar-Ilan University}
\affil[2]{UNC Chapel Hill}
}
\affil[  ]{} 
\affil[  ]{\tt \{obspp18, ramakanth.1729\}@gmail.com}
\affil[  ]{\tt \{dagan, amstery\}@cs.biu.ac.il}

\begin{document}
\maketitle
\begin{abstract}
Keyphrase extraction has been extensively researched within the single-document setting, with an abundance of methods, datasets and applications. In contrast, \textit{multi-document} keyphrase extraction has been infrequently studied, despite its utility for describing sets of documents, and its use in summarization. Moreover, no prior dataset exists for multi-document keyphrase extraction, hindering the progress of the task. Recent advances in multi-text processing make the task an even more appealing challenge to pursue. To stimulate this pursuit, we present here the first dataset for the task, \textit{\dname}, which can serve as a new benchmark, and test multiple keyphrase extraction baselines on our data. In addition, we provide a brief, yet comprehensive, literature review of the task.
\end{abstract}

\section{Introduction}
\label{sec_introduction}

Keyphrase extraction (\kpx) is the task of selecting important and topical phrases from within a body of text \citep{Turney2000LearningAF}. Single-document \kpx~has been a long standing task \citep{dennis1967keForSearch} garnering extensive research due to its vast practical uses. For example, keyphrases are listed on scientific or news articles, product descriptions and meeting transcripts to give the reader a hint at the matters of the source text. Additionally, these keyphrases are serviceable for downstream tasks like document categorization \citep{hulth2006categorization}, clustering \citep{jones2000clustering}, summarization \citep{jones2002interactive} and search \citep{gutwin1999search}.
Hence, single-document \kpx~is resourced with a multitude of datasets across several domains (e.g., scientific papers \citep{kim2009sciArticleDataset, krapivin2009sciArticleDataset} or news \citep{wan2008collabrank, marujo2012newsDataset}), and is frequently reviewed in survey papers to report on continual advancements of methods for solving the task \citep[e.g.][]{hasan2014kpexSurvey, Siddiqi2015survey, Merrouni2019review, Papagiannopoulou2020review}.

Conversely, \textit{multi-document} \kpx~(\mdkpx)~has been sporadically researched, even though it is just as valuable for indicating the central aspects of a \textit{set} of related documents. As laid out in \sect \ref{sec_review}, few works have explicitly targeted the task, however \mdkpx~was also implemented within applications of information exploration. In addition, \mdkpx~was implicitly leveraged as a medium for supporting multi-document summarization.
To make matters trickier, no dataset was previously available for \mdkpx, consisting of sets of documents and corresponding gold lists of keyphrases. Previous works, therefore, did not evaluate with standard automatic \kpx~methods, or conducted extrinsic evaluations through summarization.

To stimulate a more established research line on \mdkpx, we first briefly review the research conducted around the task (\sect \ref{sec_review}), and then present our \mdkpx~dataset,\footnote{Reproducing code will be released upon publication.}
which can provide a testing benchmark for the task (\sect \ref{sec_dataset}). The dataset is based on the existing DUC-2001 single-document \kpx~dataset \citep{wan2008collabrank} in the news domain. We leverage the properties of the original DUC-2001 multi-document \textit{summarization} dataset\footnote{\url{https://duc.nist.gov}}
to convert the single-document \kpx~dataset to a multi-document one using an automatic procedure followed by manual cleaning. We run several \kpx~algorithms on the dataset to demonstrate the current state of the task on the new benchmark (\sect \ref{sec_baselines}).

The multi-document setting poses the challenge of handling large inputs with cross-document relationships, which manifests high information redundancy along with dispersed complementing information. These issues were indeed apparent during our dataset creation process, and were treated accordingly. As for potential solutions, recent advances in multi-text processing \citep[e.g.,][]{caciularu2021cdlm, mao2020rlmmr} make \mdkpx~an even more relevant and timely task to drive forward.
\section{Task Background}
\label{sec_review}

We outline the research conducted on \mdkpx. Few works have expressly tackled the task, however it has also been applied in several studies on multi-document summarization and exploration.

\vspace{0.1cm}
\noindent
\textbf{Works on \mdkpx.}~~
Redundancy is naturally a dominant characteristic to harness for consolidating information across a set of related documents.
\citet{hammouda2005corephrase} ranked word-sequences, common to all documents, with lexical features, and evaluated resulting keyphrases against the search-query used for retrieving the set of web documents. \citet{bharti2017mdKeywordExt} also used term-frequency features, evaluating the keywords against the aggregated words in the source news articles' headlines. \citet{Qingsheng2007MDKWEx} designed cluster-based and MMR-based \citep{Carbonell1998MMR} algorithms
an
\citet{yangjie2008keyword} used TF-IDF and word-level features to score words.

Another approach taken was merging keyphrase lists from individual documents in the document-set. \citet{berend2013sdkpxForMdkpx} classified candidate keyphrases using a maxent model with features of word surface-form and Wikipedia knowledge, and unified lists with an information gain metric. The final list of keyphrases was compared to a topic overview paragraph.
\citet{bayatmakou2017queryBasedKpx} applied RAKE \citep{rose2010rake} per document and word similarity for merging. Evaluation was conducted with manual satisfaction ratings.
Relatedly, \citet{wan2008collabrank} proposed a method for \textit{single}-document \kpx, that ranks a document's keyphrases with respect to similar ``collaborating'' documents. That paper also introduced the single-document \kpx~dataset that we build upon for \mdkpx~(\sect \ref{sec_dataset}).

As apparent, the works addressing the task employ rather simplistic methods, and, notably, evaluate inconsistently and in a non-methodological manner. 
We advocate revisiting the \mdkpx~task with modern approaches, and with our dataset as a testing benchmark for comparability. While preparing \textit{training} data is left for future work, extracting keyphrases from a document set may be facilitated by semi-supervised techniques. In summarization, for example, \citet{mao2020rlmmr} used reinforcement learning against reference summaries, which can be borrowed for detecting keyphrases rather than summary sentences. \citet{lebanoff2018singleToMultiSumm} capitalized on the abundant \textit{single}-document summarization data and adapted it for the multi-document setting, as can be respectively applied in \kpx. Additionally, it is worth exploring how to leverage multi-document word representations \citep{caciularu2021cdlm} for the use of phrase salience detection.

\vspace{0.1cm}
\noindent
\textbf{Applications using \mdkpx.}~~
To alleviate the consumption of information from within document sets, there is a line of research developing interactive systems for knowledge exploration \citep{shapira2021qfse}. Many applications provide a form of a keyphrase list to highlight relevant sub-topics in the document set \citep[e.g.][]{leuski2003ineats,handler2017rookie,shapira2021qfse,hirsch2021ifacetsum}. Here too, keyphrases were extracted using redundancy-based methods, like TF-IDF, TextRank \citep{mihalcea2004textrank} or cross-document coreference resolution \citep{cattan2021cdcoref}.

\vspace{0.1cm}
\noindent
\textbf{\kpx~for multi-document summarization.}~~
Multi-document summarization (MDS) aims to generate a passage covering the salient issues of the source document-set. Keyphrases naturally point to central aspects, and can therefore assist in marking the information for a summary. Some works detected salient phrases in the document-set, e.g., with conventional term-frequency methods \citep{alshahrani2019mdsKeywordFusion}, by using single-document \kpx~algorithms on the concatenated documents \citep{nayeem2017kpexForMds}, or through query-similarity for query-focused summarization \citep{ma2008qfsWithKeywords}. \citet{hong2014MDwordImportance} assigned importance to documents' content words based on their appearance in reference summaries. ILP frameworks were also employed \citep{li2015bigramsForMDS,li2020unsupervised} for weighting phrases around which to summarize.
While most of these methods, in consequence, produce keyphrases, their intention is generating summaries that are standardly evaluated against reference summaries.





\vspace{0.1cm}
\noindent
\textbf{\kpx~evaluation.}
Most single-document \kpx~works automatically evaluate a keyphrase list against a gold list, as we now enable also for \mdkpx~with our new dataset.
The most prominent metric
is $F1@k$, which considers the recall and precision of the predicted list, truncated to $k$ items, against the full gold list. To allow for some reasonable lexical variation of keyphrases, words are often stemmed, and unigram-level $F1@k$ is used -- where the two lists of keyphrases are each flattened out to respective lists of words.

A major disadvantage of this evaluation approach is that it penalizes synonymous keyphrases not contained in the gold list. This is potentially further exacerbated in the multi-document setting, which contains higher paraphrastic diversity across documents. Our dataset annotation process facilitated preparation of \textit{substitute clusters} within gold keyphrase lists, thus allowing for some synonymy of predicted keyphrases (\sect \ref{sec_dataset}).



\section{New Dataset}
\label{sec_dataset}


Our \mdkpx~dataset, named \textit{\dname}, builds upon the DUC-2001 single-document \kpx~dataset \citep{wan2008collabrank}, for the news domain.

The DUC-2001 \textit{MDS} dataset \citep{paul2001duc01} consists of 30 topics, each containing \textasciitilde 10.3 related news articles (308 total). Experts summarized each individual article, as well as each of the document-sets, yielding
three 100-token-long summaries per document, and three summaries per document-set, at lengths 50, 100, 200 and 400 tokens.
\citet{wan2008collabrank} further annotated the data 
with lists of \textasciitilde 8.1 keyphrases per document, at \textasciitilde 2.1 words per keyphrase.
This data is still widely used for the single-document \kpx~task.

The availability of document-level keyphrases \textit{and} document clusters -- unique to the DUC-2001 dataset -- allows deducing multi-document-level keyphrases.
The single-document \kpx~dataset is restructured for the \textit{multi}-document setting by carrying out an automatic merging and reranking process, followed by a manual refinement procedure:

\vspace{0.1cm}
\noindent
\textbf{Automatic merging and reranking.}
For each topic $t$ with its corresponding document set $D_t = \{d_1, ..., d_{n_t}\}$, and 400-token reference summaries $S_t = \{s_1, s_2, s_3\}$, we first scored each stemmed word $w$ in $D_t$ as $word\_score(w, t) = avg(df(w, D_t), df(w, S_t))$ where $df(w, X)$ stands for $w$'s document-frequency in document-set $X$, i.e. the percentage of documents of $X$ in which $w$ appears. As expressed earlier, the frequency of words in the document set are useful for indicating the importance of concepts for the topic. We additionally leverage the reference summaries for providing a strong signal for topic-level salience.

We then unified $D_t$'s $n_t$ lists of keyphrases (from the single-document \kpx~dataset), removing duplicates and phrases not appearing in $D_t$,\footnote{In keyphrase \textit{extraction}, all keyphrases must be contained within the document set. We removed 16 of 2488 keyphrases that mistakenly appeared in the original single-doc dataset.} to form a single list of potential keyphrases, $K_t'$. Each phrase $p \in K_t'$ was then scored as $phrase\_score(p, t) = avg_{w \in p}(word\_score(w, t))$, i.e. the average of $p$'s stem scores. This generated a ranked list of keyphrases, $K_t$, ordered by a salience score.

Lastly, we merged pairs of phrases in $K_t$ where one was contained within the other (stemmed and disregarding word order), leaving only the longer variant or the one earlier in $K_t$, e.g., merging ``routine training''/ ``routine train flight''.
Due to the variance of keyphrases' informativeness across documents, we found that this heuristic effectively filtered out overly generic or repetitive keyphrases.

\begin{table}
    \centering
    \resizebox{\columnwidth}{!}{
    \begin{tabular}{lcc}
        \hline
         & \textit{Full} & \textit{Trunc-20} \\
        \hline
        \# topics                      & 30            & 30 \\
        Avg (StD) \# docs per topic          & 10.27 (2.24)  & 10.27 (2.24) \\
        Avg (StD) \# KPs per topic           & 43.8 (15.6)   & 19.97 (0.18) \\
        Avg (StD) KP word-length             & 2.13 (0.66)   & 2.17 (0.66) \\
        \hline
        \# KPs with substitute cluster & 142 of 1314   & 104 of 599 \\
        Avg (StD) \# KPs in clusters         & 2.82 (1.26)   & 3.07 (1.37) \\
        Avg (StD) \% unique stems in cluster & 0.72 (0.06)   & 0.71 (0.07) \\
        \hline
    \end{tabular}}
    \caption{\dname~stats, on the full data and when truncating the keyphrase lists to 20. (KP = keyphrase)}
    \label{tab_dataset_stats}
\end{table}

\vspace{0.1cm}
\noindent
\textbf{Manual refinement.}~~
As we strived to generate a high-quality \mdkpx~benchmark dataset, we further refined the keyphrase lists produced by the automatic stage above. One of the authors looked over the 30 $K_t$ lists with the relevant topic documents and reference summaries open for assistance, and carried out the following: (1) removed phrases that were particularly scarce or of low informativeness
(e.g., ``similar transmission'' in the ``Mad Cow Disease'' topic); (2) removed phrases that were not synonymous with others, but were clearly implied from other phrases (e.g., ``U.S. Senate'' where other keyphrases mention the Senate); (3) clustered together phrases that can be used replaceably (e.g., ``1990 census'' and ``1990 population count'') to form keyphrase \textit{substitute clusters}, with the more commonly used variant as the preferred alternative; (4) produced substitute clusters for persons' titled proper nouns, when the title is optional (e.g., a cluster for ``Bill Clinton'' containing ``President Clinton'' and ``Governor Bill Clinton''), leaving the untitled version as the preferred alternative. These annotation actions emphasize the need for proper consolidation of repetitive and complementing information in the multi-document setting.

The whole dataset formation procedure yielded the final \dname~dataset, with basic statistics appearing in Table \ref{tab_dataset_stats}. We suggest a version of the dataset where the keyphrase lists are truncated at 20 items, denoted here \textit{Trunc-20}. This establishes a more representational task-setting since lead keyphrases in the gold lists are more salient in their corresponding topics, while those low in the list are less anticipated as topic-level keyphrases.

Note that the variability of contextually similar keyphrases across documents enabled the formation of clusters of substitute keyphrases, which is a novel conception in \kpx~datasets. This assists in the evaluation process when a system outputs a keyphrase that is worded differently in the gold list of keyphrases -- a major shortcoming in standard \kpx~evaluation (\sect \ref{sec_review}). We marked a preferred variant in each cluster to also enable standard evaluation, that requires a flat list of gold keyphrases.

As seen in Table \ref{tab_dataset_stats}, there is a considerable amount of keyphrases that allow substitutes. Moreover, these substitute clusters are quite lexically diverse, as witnessed in the last row of the table: the percent of unique word stems within each cluster is over 70\%. In Table \ref{tab_examples} in the appendix, an example of a topic's keyphrase list clearly demonstrates the variability within clusters.

\begin{table*}
    \centering
    \resizebox{\textwidth}{!}{
    \begin{tabular}{l||cccc|cccc||cccc|cccc}
                   & \multicolumn{8}{c||}{\textbf{\textit{Concat}}} & \multicolumn{8}{c}{\textbf{\textit{Merge}}} \\\cline{2-17}
				   & \multicolumn{4}{c|}{\textbf{Precision@k}} & \multicolumn{4}{c||}{\textbf{unigram-Precision@k}} & \multicolumn{4}{c|}{\textbf{Precision@k}} & \multicolumn{4}{c}{\textbf{unigram-Precision@k}} \\
\textbf{Algorithm} & \textbf{1} & \textbf{5} & \textbf{10} & \textbf{20} & \textbf{1} & \textbf{5} & \textbf{10} & \textbf{20} & \textbf{1} & \textbf{5} & \textbf{10} & \textbf{20} & \textbf{1} & \textbf{5} & \textbf{10} & \textbf{20} \\
\hline
Tf-Idf                                              & 3.33		& 4.67		& 6.67		& 5.83		& 96.67		& 70.12		& 60.21		& 47.00		& 6.67		& 4.00		& 3.67		& 4.50		& 48.89		& 40.36		& 35.12		& 30.83 \\
KPMiner \citep{el2009kpminer}                       & 13.33		& 12.67		& 11.67		& 9.83		& 95.00		& 76.89		& 62.98		& 44.56		& 16.67		& 13.33		& 11.33		& 10.83		& 83.33		& 67.16		& 58.64		& 49.89 \\
YAKE \citep{campos2020yake}                         & 26.67		& 18.00		& 15.67		& 13.50		& 82.22		& 65.93		& 52.48		& 38.83		& 30.00		& 14.67		& 12.33		& 10.83		& 70.56		& 58.28		& 46.17		& 35.09 \\
TextRank \citep{mihalcea2004textrank}               & 6.67		& 8.67		& 7.00		& 8.67		& 54.76		& 43.07		& 34.93		& 25.23		& 26.67		& 24.67		& 20.67		& 17.17		& 77.00		& 64.80		& 55.12		& 37.80 \\
SingleRank \citep{wan2008collabrank}                & 13.33		& 13.33		& 11.00		& 12.17		& 69.06		& 48.48		& 36.92		& 27.00		& 33.33		& 22.67		& 22.67		& 18.83		& 80.33		& 64.96		& 53.78		& 39.11 \\
TopicRank \citep{bougouin2013topicrank}             & 16.67		& 14.00		& 12.00		& 10.33		& 89.44		& 63.24		& 53.84		& 42.85		& 40.00		& 29.33		& 24.00		& 18.67		& 70.00		& 67.13		& 60.03		& 45.83 \\
TopicalPageRank \citep{Sterckx2015TopiclPageRank}   & 13.33		& 14.00		& 11.67		& 13.00		& 68.58		& 49.42		& 38.33		& 27.61		& 30.00		& 26.00		& 22.33		& 20.17		& 77.00		& 65.74		& 53.44		& 40.75 \\
PositionRank \citep{florescu2017positionrank}       & 23.33		& 19.33		& 18.33		& 17.00		& 74.44		& 57.71		& 44.57		& 32.43		& 30.00		& 22.67		& 23.33		& 19.50		& 80.00		& 64.66		& 56.55		& 43.16 \\
MultipartiteRank \citep{boudin2018multipartiteRank} & 16.67		& 16.00		& 12.00		& 11.00		& 87.22		& 67.34		& 55.49		& 44.44		& 33.33		& 27.33		& 25.33		& 21.00		& 75.00		& 65.37		& 60.48		& 47.01 \\
BERT-KPE \citep{sun2021BERTKPE} 					& \multicolumn{8}{c||}{-} 																		& 23.33		& 21.33		& 21.33		& 18.50		& 74.72		& 59.89		& 49.87		& 37.30 \\	
CollabRank \citep{wan2008collabrank} 				& \multicolumn{8}{c||}{-} 																		& 30.00		& 24.00		& 22.33		& 17.83		& 80.33		& 64.50		& 52.07		& 38.48 \\	
\citep{bayatmakou2017queryBasedKpx}  [multi-doc] 	& \multicolumn{8}{c||}{-} 																		& 3.33		& 4.00		& 3.00		& 2.17		& 27.78 	& 22.24 	& 18.67		& 14.19 \\
\end{tabular}}
    \caption{Precision results on various \kpx~algorithms tested with the \textit{Trunc-20} version of our \dname~dataset. In \textit{Concat} mode all topic documents are concatenated as a single text input, and in \textit{Merge} mode algorithms are run on individual documents after which keyphrase lists are heuristically merged and reranked. The bottom two algorithms are multi-document based \kpx~algorithms, and work in \textit{Merge} mode only. BERT-KPE is limited in input size and hence cannot be run in Concat mode. This table corresponds to Table \ref{tab_baseline_results_20_f1}, which presents F1 scores.}
    \vspace{-2.5mm}
    \label{tab_baseline_results_20_prec}
\end{table*}

%


\section{Baseline Results}
\label{sec_baselines}

We demonstrate the use of \dname~by testing 11 existing single-document \kpx~algorithms and a multi-document one. Algorithms are applied in two modes: (1) \textit{Concat}, where all topic documents are concatenated into a single text that is then fed to the algorithm to output a list of keyphrases per topic; (2) \textit{Merge}, where for each topic, the algorithm is fed one document at a time, and the generated lists of keyphrases are merged using a similar strategy as in the \textit{automatic merging and reranking} procedure in \sect \ref{sec_dataset}, except that $word\_score(w, t) = df(w, D_t)$, i.e., it does not consider the reference summary set -- which is unavailable in the \kpx~task. BERT-KPE \citep{sun2021BERTKPE}, a RoBERTa \citep{liu2019roberta} model trained on the single-document OpenKP \citep{xiong2019openkp} dataset, has a strict input size limit, and cannot work in \textit{Concat} mode. CollabRank \citep{wan2008collabrank} uses its collaborating documents, hence only \textit{Merge} is applied.
The \mdkpx~algorithm by \citet{bayatmakou2017queryBasedKpx} uses a merge approach different from ours.

Table \ref{tab_baseline_results_20_prec} shows Precision@k results, with stemming, on the \textit{Trunc-20} version of \dname~and using the substitute clusters (evaluation procedure, \textit{F1} scores and scores on the full data in appendix). Overall, we witness the benefit of the \textit{Merge} strategy, which explicitly considers redundancy across documents during the merging step.
Meanwhile, some baselines tend to output many synonymous keyphrases, as seen in Table \ref{tab_baseline_stats} (appendix). On average over all baselines, \textasciitilde 15\% of output keyphrases are synonymous with others, with respect to the available substitute clusters. We may hence infer that improving detection of phrase redundancy in context, may improve overall results.
We also observe that keyphrase token-length (Table \ref{tab_baseline_stats}) influences unigram-level scores: shorter keyphrases, likely more informationally generic, tend to yield higher precision and lower recall scores.

\section{Conclusion}
\label{sec_conclusion}

We review the multi-document \kpx~task, which is far understudied compared to its single-document counterpart. While few works have tackled the \mdkpx~task head-on, without the existence of a suitable dataset, \mdkpx~has also been applied for document-set summarization and exploration.
We introduce the first \mdkpx~dataset as a benchmark, and test various \kpx~baselines on it. Alongside recent progress in multi-text processing, we hope our dataset spurs the advancement of the \mdkpx~task.

\section*{Acknowledgements}
This work was supported in part by the German Research Foundation through the German-Israeli Project Cooperation (DIP, grant DA 1600/1-1); by the Israel Science Foundation (grants no. 2827/21 and 2015/21); and by a grant from the Israel Ministry of Science and Technology.

\bibliography{bibliography}
\bibliographystyle{acl_natbib}

\clearpage
\appendix
\section{Further Experiment Details}
\label{sec_appendix_results}

\paragraph{Additional baseline evaluations.}
Table \ref{tab_baseline_results_20_f1} presents the \textit{F1} scores on the \textit{Trunc-20} version of the dataset, corresponding to Table \ref{tab_baseline_results_20_prec} in \sect \ref{sec_baselines}.
Tables \ref{tab_baseline_results_full_prec} and \ref{tab_baseline_results_full_f1} present the results on the full gold keyphrase lists (non-truncated). When compared to the results on the \textit{Trunc-20} truncated lists (Tables \ref{tab_baseline_results_20_prec} and \ref{tab_baseline_results_20_f1}), there is an expected degredation in all scores, since the keyphrases lower in the lists are less representative keyphrases of the respective document sets. This, and the longer absolute lengths of the lists, make it less likely for the \kpx~baselines to extract correct keyphrases, and hence yield considerably lower recall scores across the board (not shown here).

\paragraph{Evaluation details.}
For computing $F1@k$, the top-$k$ predicted keyphrases and all gold keyphrases are stemmed, duplicates are removed from the predicted list, and stemmed keyphrases are lexically matched. If a predicted keyphrase is found in a gold keyphrase substitute cluster, then that gold keyphrase cluster cannot be matched with another predicted keyphrase. This mimics the removal of duplicates from the predicted list, just with synonymous keyphrases. Notice that this means that repeated/synonymous keyphrases are marked as appearing only once, which affects precision unfavourably.

For computing unigram-$F1@k$: (1) the top-$k$ items in the system keyphrase list are retrieved; (2) unique keyphrases from that sub-list are flattened out to a single list of stems; (3) each substitute cluster in the gold list is flattened out to a list of unique stems; (4) all gold clusters -- including those with one element -- are pooled together to one list of stems; (5) the predicted stems are evaluated against gold stems with recall and precision.

The average precision and average \textit{F1} scores over all instances are the final scores presented.

\paragraph{\textit{Concat} mode implementation.}
When inputting one long concatenation of documents to a single-document \kpx~algorithm, the order of the documents may have an effect on the results. Therefore, for each of the 30 test topics, we shuffled the documents, and kept that order for all baselines.

\paragraph{Keyphrase sizes.}
Table \ref{tab_baseline_stats}, on the left side, presents the average token-length of the 20 keyphrases output by each baseline, over all topics, when using the \textit{Concat} and \textit{Merge} generation modes. The keyphrase sizes in \textit{Concat} are represenatative of the corresponding algorithms' output sizes, while the sizes in \textit{Merge} go through an additional process, hence slightly altering the natural output sizes of the algorithms. As mentioned in \sect \ref{sec_baselines}, the keyphrase token-length has an influence on the unigram-level precision and recall scores. When a keyphrase is shorter, it has less of a chance of containing words not in the gold keyphrases, allowing for higher precision. On the other hand, it also has less opportunity to catch those gold words, leading to lower recall.

\paragraph{KP synonymity in outputs.}
For each baseline used, Table \ref{tab_baseline_stats}, on the right side, presents the average (over 20 keyphrases per topic, and over all topics) percent of keyphrases that are synonymous with others, with respect to the substitute clusters in the test dataset (i.e., this does not take into account synonymous keyphrases that \textit{do not} appear in the gold list). Notably, we see that about 1 of every 4 keyphrases in TopicalPageRank outputs are synonymous with others. Nevertheless, this baseline is still one of the superior tested methods.

\paragraph{Single-document \kpx~results.}
We ran the algorithms from Tables \ref{tab_baseline_results_20_prec}, \ref{tab_baseline_results_20_f1}, \ref{tab_baseline_results_full_prec} and \ref{tab_baseline_results_full_f1} on the single-document DUC-2001 \kpx~dataset (308 documents and 8.08 keyphrases per document), to get a sense of their comparable quality in the single and multiple document settings. Results are presented in Table \ref{tab_sdke}. There are 7 documents that were not processed in the \textit{KPMiner} algorithm due to processing errors.

Overall, we see that the algorithm rankings are quite similar in the two settings, across the $k$ values and in both metrics.

\paragraph{Algorithm implementations.}
We used the PKE Python toolkit package \citep{boudin2016pkePackage} for most \kpx~algorithms. We adapted the code available for BERT-KPE\footnote{\url{https://github.com/thunlp/BERT-KPE}} \citep{sun2021BERTKPE} for the DUC-2001 data, and used the available trained model (RoBERTa \citep{liu2019roberta} on OpenKP data \citep{xiong2019openkp}). We implemented the algorithm by \citet{bayatmakou2017queryBasedKpx} ourselves, which uses RAKE \citep{rose2010rake} as its underlying single-document \kpx~component (we used the nltk-rake library\footnote{\url{https://pypi.org/project/rake-nltk}}). As RAKE outputted very long keyphrases yielding low scores, we used only those upto 3 words. For CollabRank, we considered all other documents in its original topic document-set as ``collaborating'' documents, and computed their similarity scores using spaCy \citep{spacy2020} text similarity.

\paragraph{Execution resources.}
All algorithms (except for BERT-KPE) and automatic methods used for annotation and experimentation were run on a standard laptop, and no special hardware was required. BERT-KPE was run (only inference was needed) on a NVIDIA GeForce GTX 1080 Ti GPU with 11GB memory, and used less than 1GB memory during inference.

Run times were upto about a second per keyphrase extraction instance, except for CollabRank which required about 15-20 seconds per document. Running the \textit{Merge} mode on the document-sets required tens of seconds for some baselines as the process iterates over all documents separately. The \textit{Concat} mode, which requires a single run per document-set, was substantially faster overall.

\section{Dataset}
\label{sec_appendix_example}

\paragraph{Distribution of data.}
For our work, the DUC-2001 MDS dataset was obtained according to NIST instructions, and the DUC-2001 keyphrases were taken from \url{github.com/boudinfl/duc-2001-pre}. Since the documents from the DUC-2001 dataset cannot be freely re-distributed, we make available a script for one-click \textit{\dname}~dataset re-construction using the properly acquired DUC-2001 MDS dataset from NIST.

\paragraph{Example.}
Table \ref{tab_examples} presents an example list of keyphrases from our \dname~dataset. The top 20 keyphrases are used in the \textit{Trunc-20} dataset version, while the full list is used in the full dataset version. Some keyphrases have multiple wording variations, acting as the \textit{substitute clusters}. The first item in a cluster can be used in the standard evaluation when a flat list of keyphrases is required.

\section{Previous \mdkpx{} Evaluation Methods}
\label{sec_appendix_previous_evaluations}

As discussed in \sect \ref{sec_review}, the previous works explicitly solving the \mdkpx{} task did not have a proper dataset to test their resulting keyphrases. Consequently, each work tested their results differently, described as follows:

\citet{hammouda2005corephrase} targeted the web-document domain. To evaluate, 10 sets of ~30 documents were retrieved via query search by submitting a short query (2 to 3 words) into a search engine for each such set (about 30 documents per set with 500 words per document). The system keyphrases were compared, by word-stem overlap, to the \textit{single} corresponding document-set search query, as an indicator for keyphrase-salience.

\citet{berend2013sdkpxForMdkpx} work on the scientific paper domain. Sets of papers from ACL workshops \citep{schafer2012workshopSet} focusing on a clearly distinguishable scientific area (110 workshops with \textasciitilde{14} articles each) were paired with their respective ``call-for-papers'' (CFP) website sections. A system keyphrase list on a paper set was then compared to the CFP text via word-level cosine similarity. Also, NLP experts assessed whether keyphrase lists indeed properly characterized the corresponding workshop.

\citet{bayatmakou2017queryBasedKpx} retrieved common documents with a search query. While an automatic evaluation was proposed (measuring against the search query, and co-occurrence of keywords and query in documents), the actual assessment was a manual satisfaction rating against the search query. Experiments were performed over a large dataset of $13,870$ scientific abstracts (\url{https://www.webofknowledge.com}).

\citet{bharti2017mdKeywordExt} evaluated the resulting keyword list against the aggregated words in the news articles' headlines, with recall and precision.

\citet{Qingsheng2007MDKWEx} tested against proprietary expert-annotated data.

\begin{table*}
    \centering
    \resizebox{\textwidth}{!}{
    \begin{tabular}{l||cccc|cccc||cccc|cccc}
                   & \multicolumn{8}{c||}{\textbf{\textit{Concat}}}                                 & \multicolumn{8}{c}{\textbf{\textit{Merge}}}                         \\\cline{2-17}
				   & \multicolumn{4}{c|}{\textbf{F1@k}}                                 & \multicolumn{4}{c||}{\textbf{unigram-F1@k}}                         & \multicolumn{4}{c|}{\textbf{F1@k}}                                 & \multicolumn{4}{c}{\textbf{unigram-F1@k}}                         \\
\textbf{Algorithm} & \textbf{1} & \textbf{5} & \textbf{10} & \textbf{20} & \textbf{1} & \textbf{5} & \textbf{10} & \textbf{20} & \textbf{1} & \textbf{5} & \textbf{10} & \textbf{20} & \textbf{1} & \textbf{5} & \textbf{10} & \textbf{20} \\
\hline
Tf-Idf                                              & 0.32		& 1.87		& 4.44		& 5.83		& 4.56		& 16.92		& 26.53		& 34.84     & 0.63		& 1.60		& 2.44		& 4.50		& 4.52		& 17.04		& 24.04		& 31.00		\\
KPMiner \citep{el2009kpminer}                       & 1.27		& 5.07		& 7.79		& 9.84		& 5.52		& 21.56		& 31.30		& 35.16     & 1.59		& 5.34		& 7.56		& 10.84		& 5.39		& 19.45		& 28.85		& 38.93		\\
YAKE \citep{campos2020yake}                         & 2.54		& 7.20		& 10.44		& 13.50		& 6.24		& 24.67		& 31.99		& 36.33     & 2.86		& 5.87		& 8.23		& 10.84		& 7.06		& 25.75		& 33.87		& 37.72		\\
TextRank \citep{mihalcea2004textrank}               & 0.63		& 3.47		& 4.67		& 8.68		& 10.81		& 27.90		& 32.72		& 31.21     & 2.54		& 9.88		& 13.79		& 17.17		& 9.21		& 28.79		& 41.19		& 41.66		\\
SingleRank \citep{wan2008collabrank}                & 1.27		& 5.34		& 7.35		& 12.18		& 12.67		& 29.98		& 32.93		& 32.02     & 3.17		& 9.07		& 15.12		& 18.84		& 9.52		& 28.56		& 39.33		& 42.18		\\
TopicRank \citep{bougouin2013topicrank}             & 1.59		& 5.60		& 8.00		& 10.33		& 5.36		& 18.55		& 27.70		& 35.28     & 3.81		& 11.74		& 16.01		& 18.68		& 6.88		& 25.36		& 39.01		& 44.84		\\
TopicalPageRank \citep{Sterckx2015TopiclPageRank}   & 1.27		& 5.61		& 7.79		& 13.01		& 12.29		& 29.77		& 33.57		& 32.32     & 2.86		& 10.41		& 14.90		& 20.18		& 9.27		& 28.77		& 38.88		& 43.48		\\
PositionRank \citep{florescu2017positionrank}       & 2.22		& 7.74		& 12.24		& 17.01		& 9.44		& 27.76		& 33.75		& 34.75     & 2.86		& 9.07		& 15.56		& 19.51		& 8.27		& 27.28		& 39.15		& 44.47		\\
MultipartiteRank \citep{boudin2018multipartiteRank} & 1.59		& 6.40		& 8.00		& 11.00		& 5.62		& 20.35		& 28.75		& 36.55     & 3.17		& 10.94		& 16.90		& 21.01		& 7.45		& 25.14		& 39.28		& 46.08		\\
BERT-KPE \citep{sun2021BERTKPE} 					& \multicolumn{8}{c||}{-} 																		& 2.22		& 8.54		& 14.23		& 18.51		& 8.06		& 28.53		& 38.04		& 41.56		\\
CollabRank \citep{wan2008collabrank} 				& \multicolumn{8}{c||}{-}																		& 2.86		& 9.61		& 14.90		& 17.84		& 9.37		& 28.03		& 37.68		& 41.26		\\
\citep{bayatmakou2017queryBasedKpx}  [multi-doc] 	& \multicolumn{8}{c||}{-}                                                                       & 0.32		& 1.60		& 2.00		& 2.17		& 3.57		& 11.29		& 15.04		& 16.31		\\
\end{tabular}}
    \caption{F1 results on various \kpx~algorithms tested with the \textit{Trunc-20} version of our \dname~dataset. In \textit{Concat} mode all topic documents are concatenated as a single text input, and in \textit{Merge} mode algorithms are run on individual documents after which keyphrase lists are heuristically merged and reranked. The bottom two algorithms are multi-document based \kpx~algorithms, and work in \textit{Merge} mode only. BERT-KPE is limited in input size and hence cannot be run in Concat mode. This table corresponds to Table \ref{tab_baseline_results_20_prec}, which presents precision scores.}
    \label{tab_baseline_results_20_f1}
\end{table*}
\begin{table*}
    \centering
    \resizebox{\textwidth}{!}{
    \begin{tabular}{l||cccc|cccc||cccc|cccc}
                   & \multicolumn{8}{c||}{\textbf{\textit{Concat}}}                                 & \multicolumn{8}{c}{\textbf{\textit{Merge}}}                         \\\cline{2-17}
				   & \multicolumn{4}{c|}{\textbf{Precision@k}}                                 & \multicolumn{4}{c||}{\textbf{unigram-Precision@k}}                         & \multicolumn{4}{c|}{\textbf{Precision@k}}                                 & \multicolumn{4}{c}{\textbf{unigram-Precision@k}}                         \\
\textbf{Algorithm} & \textbf{1} & \textbf{5} & \textbf{10} & \textbf{20} & \textbf{1} & \textbf{5} & \textbf{10} & \textbf{20} & \textbf{1} & \textbf{5} & \textbf{10} & \textbf{20} & \textbf{1} & \textbf{5} & \textbf{10} & \textbf{20} \\
\hline
Tf-Idf												& 3.33		& 5.33		& 7.33		& 6.50		& 100.0		& 74.26		& 65.55		& 53.61		& 6.67		& 4.00		& 3.67		& 4.50		& 50.56		& 42.66		& 37.46		& 34.29		\\
KPMiner \citep{el2009kpminer}                       & 13.33		& 13.33		& 12.67		& 10.83		& 98.33		& 81.19		& 68.29		& 52.42		& 16.67		& 13.33		& 11.33		& 11.67		& 86.67		& 69.59		& 61.94		& 55.52		\\
YAKE \citep{campos2020yake}                         & 30.00		& 18.67		& 17.00		& 14.33		& 88.89		& 72.33		& 59.64		& 44.84		& 30.00		& 14.67		& 12.33		& 11.17		& 72.78		& 64.61		& 51.54		& 40.77		\\
TextRank \citep{mihalcea2004textrank}               & 6.67		& 10.00		& 7.67		& 9.83		& 63.58		& 50.02		& 42.41		& 33.65		& 26.67		& 24.67		& 20.67		& 18.17		& 80.33		& 72.00		& 62.02		& 45.92		\\
SingleRank \citep{wan2008collabrank}                & 13.33		& 14.00		& 12.00		& 13.17		& 73.73		& 55.06		& 43.73		& 34.32		& 33.33		& 22.67		& 22.67		& 19.33		& 80.33		& 71.18		& 60.67		& 47.08		\\
TopicRank \citep{bougouin2013topicrank}             & 16.67		& 16.00		& 13.67		& 12.00		& 92.78		& 67.75		& 59.52		& 52.87		& 40.00		& 29.33		& 24.33		& 20.00		& 70.00		& 73.60		& 65.73		& 53.41		\\
TopicalPageRank \citep{Sterckx2015TopiclPageRank}   & 13.33		& 15.33		& 13.33		& 14.33		& 76.42		& 56.86		& 46.40		& 35.30		& 30.00		& 26.00		& 22.33		& 20.33		& 80.33		& 72.55		& 59.93		& 48.26		\\
PositionRank \citep{florescu2017positionrank}       & 30.00		& 22.00		& 21.00		& 19.00		& 86.67		& 65.49		& 52.08		& 40.68		& 30.00		& 22.67		& 23.67		& 20.00		& 85.00		& 69.78		& 63.26		& 50.54		\\
MultipartiteRank \citep{boudin2018multipartiteRank} & 16.67		& 18.67		& 13.67		& 13.33		& 90.56		& 73.11		& 61.56		& 53.59		& 33.33		& 28.00		& 25.67		& 22.67		& 75.00		& 72.27		& 65.65		& 54.08		\\
BERT-KPE \citep{sun2021BERTKPE} 					& \multicolumn{8}{c||}{-}																		& 23.33		& 22.67		& 22.67		& 20.67		& 76.67		& 65.83		& 57.43		& 45.50		\\
CollabRank \citep{wan2008collabrank} 				& \multicolumn{8}{c||}{-}                                                                       & 30.00		& 24.00		& 22.33		& 18.17		& 83.67		& 71.74		& 58.96		& 46.68		\\
\citep{bayatmakou2017queryBasedKpx}  [multi-doc] 	& \multicolumn{8}{c||}{-}                                                                       & 3.33		& 4.00		& 3.00		& 2.17		& 27.78		& 22.24		& 18.67		& 14.19		\\
\end{tabular}}
    \caption{Precision results on various \kpx~algorithms tested with our full \dname~dataset. In \textit{Concat} mode all topic documents are concatenated as a single text input, and in \textit{Merge} mode algorithms are run on individual documents after which keyphrase lists are heuristically merged and reranked. The bottom two algorithms are multi-document based \kpx~algorithms, and work in \textit{Merge} mode only. BERT-KPE is limited in input size and hence cannot be run in Concat mode. This table corresponds to Table \ref{tab_baseline_results_full_f1}, which presents F1 scores.}
    \label{tab_baseline_results_full_prec}
\end{table*}

\begin{table*}
    \centering
    \resizebox{\textwidth}{!}{
    \begin{tabular}{l||cccc|cccc||cccc|cccc}
                   & \multicolumn{8}{c||}{\textbf{\textit{Concat}}}                                 & \multicolumn{8}{c}{\textbf{\textit{Merge}}}                         \\\cline{2-17}
				   & \multicolumn{4}{c|}{\textbf{F1@k}}                                 & \multicolumn{4}{c||}{\textbf{unigram-F1@k}}                         & \multicolumn{4}{c|}{\textbf{F1@k}}                                 & \multicolumn{4}{c}{\textbf{unigram-F1@k}}                         \\
\textbf{Algorithm} & \textbf{1} & \textbf{5} & \textbf{10} & \textbf{20} & \textbf{1} & \textbf{5} & \textbf{10} & \textbf{20} & \textbf{1} & \textbf{5} & \textbf{10} & \textbf{20} & \textbf{1} & \textbf{5} & \textbf{10} & \textbf{20} \\
\hline
Tf-Idf												& 0.11		& 1.22		& 2.91		& 4.37		& 2.62		& 9.99		& 16.62		& 24.38     & 0.54		& 1.21		& 1.67		& 3.33		& 2.58		& 10.58		& 15.59		& 22.59     \\
KPMiner \citep{el2009kpminer}                       & 0.95		& 3.14		& 5.32		& 7.44		& 3.26		& 12.85		& 19.73		& 25.56     & 1.06		& 3.29		& 5.04		& 7.90		& 3.11		& 11.31		& 18.33		& 27.32     \\
YAKE \citep{campos2020yake}                         & 1.58		& 4.48		& 6.99		& 9.73		& 3.66		& 15.16		& 21.58		& 27.15     & 1.54		& 3.37		& 5.12		& 7.52		& 4.13		& 16.14		& 23.16		& 29.56     \\
TextRank \citep{mihalcea2004textrank}               & 0.45		& 2.48		& 3.19		& 6.73		& 7.15		& 19.92		& 25.91		& 29.39     & 1.36		& 5.45		& 8.04		& 11.60		& 5.27		& 18.32		& 28.42		& 33.92     \\
SingleRank \citep{wan2008collabrank}                & 0.73		& 3.48		& 5.04		& 8.98		& 7.83		& 20.74		& 25.16		& 28.36     & 1.61		& 4.82		& 8.83		& 12.36		& 5.27		& 17.76		& 27.12		& 33.65     \\
TopicRank \citep{bougouin2013topicrank}             & 0.86		& 3.65		& 5.37		& 8.09		& 3.08		& 11.48		& 18.07		& 27.54     & 2.18		& 6.95		& 9.92		& 13.34		& 4.04		& 16.23		& 26.15		& 34.32     \\
TopicalPageRank \citep{Sterckx2015TopiclPageRank}   & 0.73		& 3.71		& 5.56		& 9.67		& 7.86		& 20.70		& 25.96		& 28.45     & 1.53		& 5.69		& 8.78		& 13.31		& 5.44		& 18.2		& 26.65		& 34.08     \\
PositionRank \citep{florescu2017positionrank}       & 1.47		& 4.94		& 8.31		& 12.54		& 5.80		& 18.26		& 24.08		& 28.74     & 1.52		& 4.84		& 9.29		& 13.10		& 4.93		& 16.8		& 26.40		& 34.31     \\
MultipartiteRank \citep{boudin2018multipartiteRank} & 0.86		& 4.05		& 5.57		& 8.79		& 3.27		& 12.46		& 18.80		& 27.75     & 1.86		& 6.68		& 10.42		& 15.11		& 4.50		& 16.2		& 26.26		& 34.92     \\
BERT-KPE \citep{sun2021BERTKPE} 					& \multicolumn{8}{c||}{-}																		& 1.28		& 5.17		& 8.93		& 13.20		& 4.57		& 18.00		& 26.72		& 33.85     \\
CollabRank \citep{wan2008collabrank} 				& \multicolumn{8}{c||}{-}                                                                       & 1.56		& 5.37		& 8.80		& 11.89		& 5.28		& 17.83		& 25.82		& 33.20     \\
\citep{bayatmakou2017queryBasedKpx}  [multi-doc] 	& \multicolumn{8}{c||}{-}                                                                       & 0.32		& 1.60		& 2.00		& 2.17		& 3.57		& 11.29		& 15.04		& 16.31     \\
\end{tabular}}
    \caption{F1 results on various \kpx~algorithms tested with our full \dname~dataset. In \textit{Concat} mode all topic documents are concatenated as a single text input, and in \textit{Merge} mode algorithms are run on individual documents after which keyphrase lists are heuristically merged and reranked. The bottom two algorithms are multi-document based \kpx~algorithms, and work in \textit{Merge} mode only. BERT-KPE is limited in input size and hence cannot be run in Concat mode. This table corresponds to Table \ref{tab_baseline_results_full_prec}, which presents precision scores.}
    \label{tab_baseline_results_full_f1}
\end{table*}

\begin{table*}[t]
    \centering
    \begin{tabular}{l||c|c||c|c}
        \hline
                          & \multicolumn{2}{c||}{Avg. KP Word Count} & \multicolumn{2}{c}{Avg. \% Synon. KPs} \\
        Algorithm         & \textit{Concat} & \textit{Merge}    & \textit{Concat} & \textit{Merge} \\
        \hline
        Tf-Idf            & 1.30            & 2.22              & 2                 & 7     \\
        KPMiner           & 1.42            & 1.39              & 4                 & 1     \\
        YAKE              & 1.99            & 2.58              & 9                 & 17    \\
        TextRank          & 3.64            & 2.68              & 18                & 21    \\
        SingleRank        & 3.24            & 2.57              & 22                & 23    \\
        TopicRank         & 1.51            & 2.08              & 4                 & 13    \\
        TopicalPageRank   & 3.14            & 2.52              & 27                & 26    \\
        PositionRank      & 2.52            & 2.32              & 26                & 25    \\
        MultipartiteRank  & 1.51            & 2.12              & 5                 & 14    \\
        BERT-KPE          & -               & 2.81              & -                 & 19    \\
        CollabRank        & -               & 2.54              & -                 & 23    \\
        \citep{bayatmakou2017queryBasedKpx} & - & 3.00          & -                 & 0     \\
        \hline
    \end{tabular}
    \caption{The average (over all topics) number of tokens per keyphrase produced by the different algorithms, and the average (over all topics) percent of keyphrases in a topic that are ``synonymous'', i.e., share substitute clusters with others, in the \textit{Trunc-20} dataset version. Results are shown for the two generation modes (\textit{Concat} and \textit{Merge}), on the 20 output keyphrases of each baselines.}
    \label{tab_baseline_stats}
\end{table*}
\begin{table*}
    \centering
    \resizebox{\textwidth}{!}{
    \begin{tabular}{l||cc|cc|cc|cc||cc|cc|cc|cc||c}
    \hline
                   & \multicolumn{8}{c||}{\textbf{Precision/F1@k}}                  & \multicolumn{8}{c||}{\textbf{unigram-Precision/F1@k}}         & \textbf{Avg. KP} \\
\textbf{Algorithm} & \multicolumn{2}{c|}{\textbf{1}} & \multicolumn{2}{c|}{\textbf{5}} & \multicolumn{2}{c|}{\textbf{10}} & \multicolumn{2}{c||}{\textbf{20}} & \multicolumn{2}{c|}{\textbf{1}} & \multicolumn{2}{c|}{\textbf{5}} & \multicolumn{2}{c|}{\textbf{10}} & \multicolumn{2}{c||}{\textbf{20}} & \textbf{Length}  \\
\hline                                                                                                                                     
Tf-Idf            & 14.61  & 3.45  & 11.75  & 9.28   & 9.74   & 11.02  & 7.66   & 11.02  & 86.53   & 11.12  & 54.16  & 29.46  & 38.17    & 33.64  & 24.77    & 30.99  & 1.50 \\
KPMiner          & 27.24  & 6.43  & 18.18  & 14.31  & 12.55  & 13.98  & 8.31   & 11.40  & 81.73   & 14.15  & 57.51  & 34.06  & 44.52    & 37.82  & 34.93    & 37.61  & 1.17 \\
YAKE             & 18.83  & 4.43  & 15.84  & 12.44  & 12.44  & 13.85  & 9.87   & 14.09  & 54.65   & 13.04  & 37.11  & 28.55  & 26.72    & 29.17  & 19.73    & 27.41  & 2.02 \\
TextRank         & 12.99  & 3.16  & 14.81  & 11.47  & 15.88  & 17.49  & 14.04  & 19.54  & 40.51   & 15.93  & 29.68  & 29.39  & 24.12    & 30.30  & 18.71    & 27.80  & 2.85 \\
SingleRank       & 29.55  & 7.04  & 25.72  & 20.06  & 21.33  & 23.56  & 17.09  & 23.93  & 61.70   & 21.20  & 38.03  & 35.41  & 28.45    & 34.55  & 21.19    & 30.95  & 2.59 \\
TopicRank        & 35.39  & 8.36  & 25.01  & 19.51  & 19.41  & 21.48  & 14.09  & 19.60  & 75.97   & 14.86  & 53.54  & 35.54  & 40.93    & 39.89  & 29.25    & 37.28  & 1.58 \\
TopicalPageRank  & 32.14  & 7.59  & 26.89  & 20.98  & 22.26  & 24.58  & 17.82  & 24.93  & 63.36   & 21.40  & 39.46  & 36.04  & 29.46    & 35.30  & 22.15    & 32.05  & 2.50   \\
PositionRank     & 36.04  & 8.22  & 29.65  & 22.97  & 25.19  & 27.76  & 19.16  & 26.77  & 63.42   & 17.56  & 43.23  & 35.42  & 34.48    & 38.38  & 25.30 & 35.09  & 2.08 \\
MultipartiteRank & 38.96  & 9.16  & 27.41  & 21.42  & 21.20  & 23.51  & 15.94  & 22.28  & 77.65   & 15.65  & 55.56  & 36.83  & 43.73    & 42.49  & 31.74    & 40.37  & 1.56 \\
BERT-KPE          & 41.56  & 9.66  & 29.55  & 23.14  & 22.66  & 25.32  & 16.20 & 23.08  & 63.43   & 14.10  & 46.89  & 34.81  & 35.02    & 37.94  & 23.32    & 32.74  & 2.10 \\
CollabRank		 & 37.99  & 8.94  &	29.23  & 22.88  & 24.29	 & 26.92  & 18.18  & 25.50  & 70.65   & 22.37  & 42.10  & 36.92  & 31.68    & 37.38  & 23.16    & 33.31  & 2.41 \\
\hline
\end{tabular}}
    \caption{The results of various single-document \kpx~algorithms on the \textit{single}-document DUC-2001 \kpx~dataset \citep{wan2008collabrank}, for reference as a comparison to algorithms' results in the \textit{multi}-document setting (Tables \ref{tab_baseline_results_20_prec}, \ref{tab_baseline_results_20_f1}, \ref{tab_baseline_results_full_prec}, \ref{tab_baseline_results_full_f1} and \ref{tab_baseline_stats}). The average number of KPs in each document's gold list in the dataset is 8.08, and all KPs are used in the evaluation. CollabRank is a single-document \kpx~algorithm that uses related documents (within the same topic) in its operation.}
    \label{tab_sdke}
\end{table*}

\begin{table*}
    \centering
    \small
    \begin{tabular}{|c|l|}
        \hline
        \textbf{\#}         & \textbf{Keyphrase} \\
		\hline
        1                   & drug testing \\
        \hline
        \multirow{3}{*}{2}  & illegal steroid use \\
                            & drug use \\
                            & illegal performance-enhancing drugs \\
        \hline
        3                   & Olympics gold medal \\
        \hline
        4                   & Seoul Olympics \\
        \hline
        \multirow{2}{*}{5}  & banned steroid \\
                            & illegal anabolic steroid \\
        \hline
        \multirow{4}{*}{6}  & Ben Johnson \\
                            & Canadian Ben Johnson \\
                            & Sprinter Ben Johnson \\
                            & Canadian Olympic sprinter \\
        \hline
        7                   & world record \\
        \hline
        \multirow{2}{*}{8}  & anabolic steroid stanozolol \\
                            & illegal steroid stanzolol \\
        \hline
        9                   & world championships \\
        \hline
        \multirow{3}{*}{10} & Charlie Francis \\
                            & Canadian coach Charlie Francis \\
                            & Canadian national sprint coach \\
        \hline
        \multirow{2}{*}{11} & 100-meter dash \\
                            & 100-metre sprint \\
        \hline
        12                  & stanozolol use \\
        \hline
        \multirow{3}{*}{13} & Carl Lewis \\
                            & American Carl Lewis \\
                            & U.S. sprinter Carl Lewis \\
        \hline
        14                  & urine sample \\
        \hline
        15                  & steroid furazabol \\
        \hline
        16                  & Jamie Astaphan \\
        \hline
        17                  & steroid combination \\
        \hline
        18                  & Toronto \\
        \hline
        19                  & personal physician \\
        \hline
        20                  & disgraced Olympic sprinter \\
        \hline \hline
        \multirow{2}{*}{21} & Canadian inquiry \\
                            & federal inquiry \\
        \hline
        22                  & drug scandal \\
        \hline
        23                  & Angella Issajenko \\
        \hline
        24                  & Johnson scandal \\
        \hline
        25                  & stripping \\
        \hline
        26                  & controlled substance \\
        \hline
        27                  & world record-holder \\
        \hline
        28                  & Hamilton spectator indoor games \\
        \hline
        29                  & disappointed nation \\
        \hline
        30                  & record crowd \\
        \hline
        31                  & world-class sprinter \\
        \hline
        32                  & two-year suspension \\
        \hline
        33                  & news conference \\
        \hline
        34                  & first race \\
        \hline
        35                  & second-place finish \\
        \hline
        36                  & Lynda Huey \\
        \hline
        37                  & first indoor loss \\
        \hline
        38                  & slow start \\
        \hline
        39                  & Daron Council \\
        \hline
        40                  & homecoming \\
        \hline
        41                  & expectation \\
		\hline
    \end{tabular}
    \caption{The keyphrases in our \dname~dataset for topic d31 about the Ben Johnson steroid scandal, containing 13 documents. Keyphrases with multiple items represent \textit{substitute clusters}, where the first item in the cluster is the marked preferred keyphrase wording when using standard \kpx~evaluation using a flat list of gold keyphrases. The top 20 keyphrases are used in the \textit{Trunc-20} dataset version.}
    \label{tab_examples}
\end{table*}

\end{document}